\documentclass{article} 
\usepackage{nips15submit_e,times}
\usepackage{hyperref}
\usepackage{url}
\usepackage{authblk}
\usepackage{amsmath,graphicx,booktabs}
\usepackage{epstopdf,amssymb}
\usepackage{multirow}
\usepackage{fixltx2e}
\usepackage{subfigure}
\usepackage{float}
\usepackage{setspace}

\title{\pmb{Efforts estimation of doctors annotating medical image}} 

\author{Yang Deng$^{1,2}$, Yao Sun$^{1,2\#}$, Yongpei Zhu$^{1,2}$, Yue Xu$^{1,2}$, Qianxi Yang$^{1,2}$, Shuo Zhang$^{2}$, Mingwang Zhu$^{3}$, Jirang Sun$^{3}$, Weiling Zhao$^{4}$, Xiaobo Zhou$^{4*}$, Kehong Yuan$^{1*}$\\
	$^{1}$Graduate School at Shenzhen, Tsinghua University, Shenzhen 518055, China.\\
	$^{2}$Department of Biomedical Engineering, Tsinghua University, Beijing 100084, China.\\
	$^{3}$Beijing Sanbo Brain Hospital, Beijing 100825, China.\\
	$^{4}$School of Biomedical Informatics,The University of Texas Health Science Center at Houston,USA\\
	$^{\#}$Yang Deng and Yao Sun contributed equally\\
	*Corresponding author:Xiaobo Zhou(e-mail:zhouxb2015@163.com)\\ 
	and Kehong Yuan (e-mail:yuankh@sz.tsinghua.edu.cn)
}

\nipsfinalcopy 

\begin{document}
\begin{spacing}{1.2}

\maketitle

\begin{abstract}
Accurate annotation of medical image is the crucial step for image AI clinical application. However, annotating medical image will incur a great deal of annotation effort and expense due to its high complexity and needing experienced doctors. To alleviate annotation cost, some active learning methods are proposed. But such methods just cut the number of annotation candidates and don’t study how many efforts the doctor will exactly take, which is not enough since even annotating a small amount of medical data will take a lot of time for the doctor. \\

In this paper, we propose a new criterion to evaluate efforts of doctors annotating medical image. First, by coming active learning and U-shape network, we employ a suggestive annotation strategy to choose the most effective annotation candidates. Then we exploit a fine annotation platform to alleviate annotating efforts on each candidate and first utilize a new criterion to quantitatively calculate the efforts taken by doctors. In our work, we take MR brain tissue segmentation as an example to evaluate the proposed method.\\

Extensive experiments on the well-known IBSR18 dataset and MRBrainS18 Challenge dataset show that, using proposed strategy, state-of-the-art segmentation performance can be achieved by using only 60\% annotation candidates and annotation efforts can be alleviated by at least 44\%, 44\%, 47\% on CSF, GM, WM separately.\\

\pmb{Keywords}: medical image segmentation, MRI, active learning, annotation efforts
\end{abstract}

\section{Introduction}
Medical image segmentation is the base for diagnosis, surgical planning, and treatment of diseases. Recent advances in deep learning\cite{Bao2015Multi}\cite{Chen2017VoxResNet}\cite{Hao2016Deep}\cite{Hao2016DCAN}\cite{Xu2016Gland}\cite{Moeskops2016Automatic}\cite{Dong2016Fully}\cite{Zhang2015Deep}have achieved promising results on many biomedical image segmentation tasks. Relying on large annotated datasets, deep learning can achieve promising results. However, differing from natural scene images, labeled medical data are too rare and expensive to extensively obtain since annotating medical image is not only tedious and time consuming but also can only be effectively performed by medical experts.

To dramatically alleviate the common burden of manual annotation, some weakly supervised segmentation algorithms\cite{Hong2015Decoupled}and active learning\cite{google}\cite{Wang2016A}\cite{Jain2016Active}have been proposed. However, these methods are used in natural scene image analysis, which cannot be easily imitated in biomedical image settings due to large variations and rare training data in biomedical applications. For biomedical images, Zhou et al.\cite{Zhou2017Fine}presented fine-tuning convolutional neural networks for colonoscopy frame classification, polyp detection, and pulmonary embolism (PE) detection. Lin et al.\cite{Yang}presented an annotation suggestion for lymph node ultrasound image and gland segmentation by combining FCNs and active learning. But such methods just cut the number of annotation candidates, which is not enough for medical image since even annotating a small amount of data will take a lot of time for the doctor. As is shown in Fig.1, for example, annotating MR image is difficult, which will take more than 20 hours by a doctor to annotate a set of volume MR size of   and seldom experts are willing to do it, for the high complex structure and little grayscale change between different tissue classes of MR images.\\

\begin{figure}[H]
	\begin{center}
		\includegraphics[width=1.\linewidth]{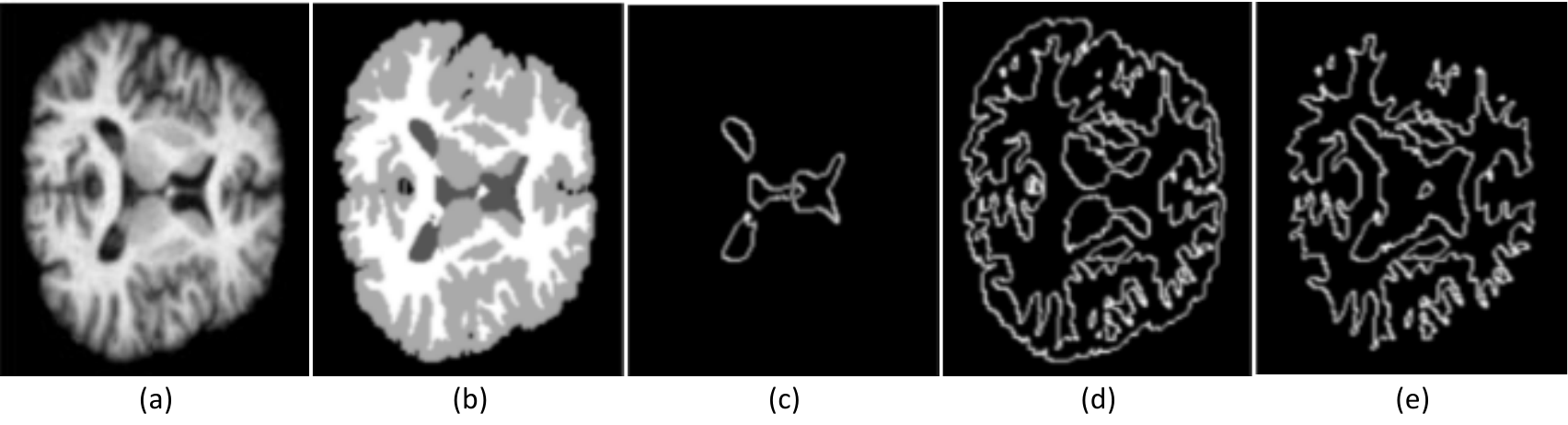}
		\caption{(a) an original image; (b) complete ground truth; (c) ground truth of cerebrospinal fluid (CSF); (d) ground truth of grey matter (GM); (e) ground truth of white matter (WM).}
	\end{center}
\end{figure}

In this paper, we propose a new criterion to evaluate efforts of doctors annotating medical image. There are two major components: (1) suggestive annotation to reduce annotation candidates; (2) an annotation platform of fine annotation to alleviate annotating efforts on each candidate. We take MR brain tissue segmentation as an example to evaluate proposed method. Extensive experiments using the well-known IBSR18 dataset\footnote{\url{https://www.nitrc.org/frs/?group_id=48}} and MRBrainS18 Challenge dataset\footnote{\url{https://mrbrains18.isi.uu.nl/data/}} show our proposed method can attain state-of-the-art segmentation performance by using only 60\% training data and annotation efforts will be cut separately at least 44\%, 44\%, 47\% for CSF,GM,WM on each annotation candidate. The remainder of this paper is organized as follows. In section 2, we introduce proposed method. Experiments and results are detailed in section 3. Finally, the discussion and the main conclusions are illustrated in section 4 and section 5 separately.

\section{Method}
We first exploit suggestive annotation to reduce annotation candidates. Then we employ our annotation platform to alleviate annotation efforts on each candidate, which are quantitatively calculated by proposed criterion.
\begin{figure}[H]
	\begin{center}
		\includegraphics[width=1.\linewidth]{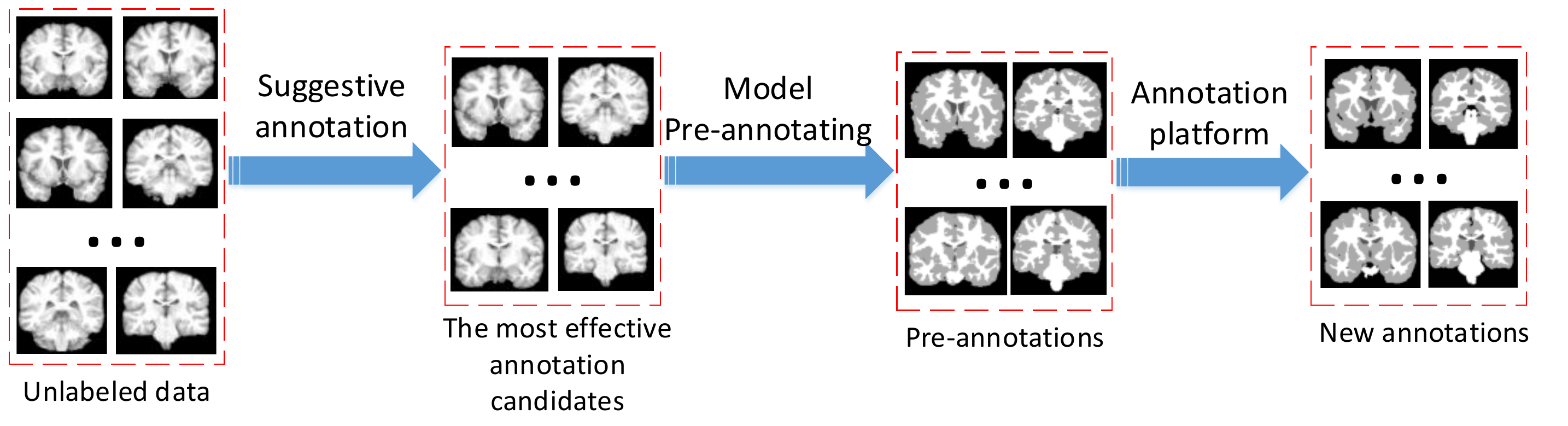}
		\caption{illustrating the overall of the proposed method.}
	\end{center}
\end{figure}

\subsection{Suggestive annotation}
We present the annotation strategy by combining U-shape network model and active learning. Fig.3. illustrates the main ideas and steps of proposed strategy. Starting with very little training data, we iteratively train a set of U-shape models. At the end of each stage, if the test results cannot meet the requirements of experts, we extract uncertainty estimation from these models to decide what will be the next data to annotate. After acquiring the new annotation data, the next stage is started using all available annotated data.

\begin{figure}[H]
	\begin{center}
		\includegraphics[width=1.\linewidth]{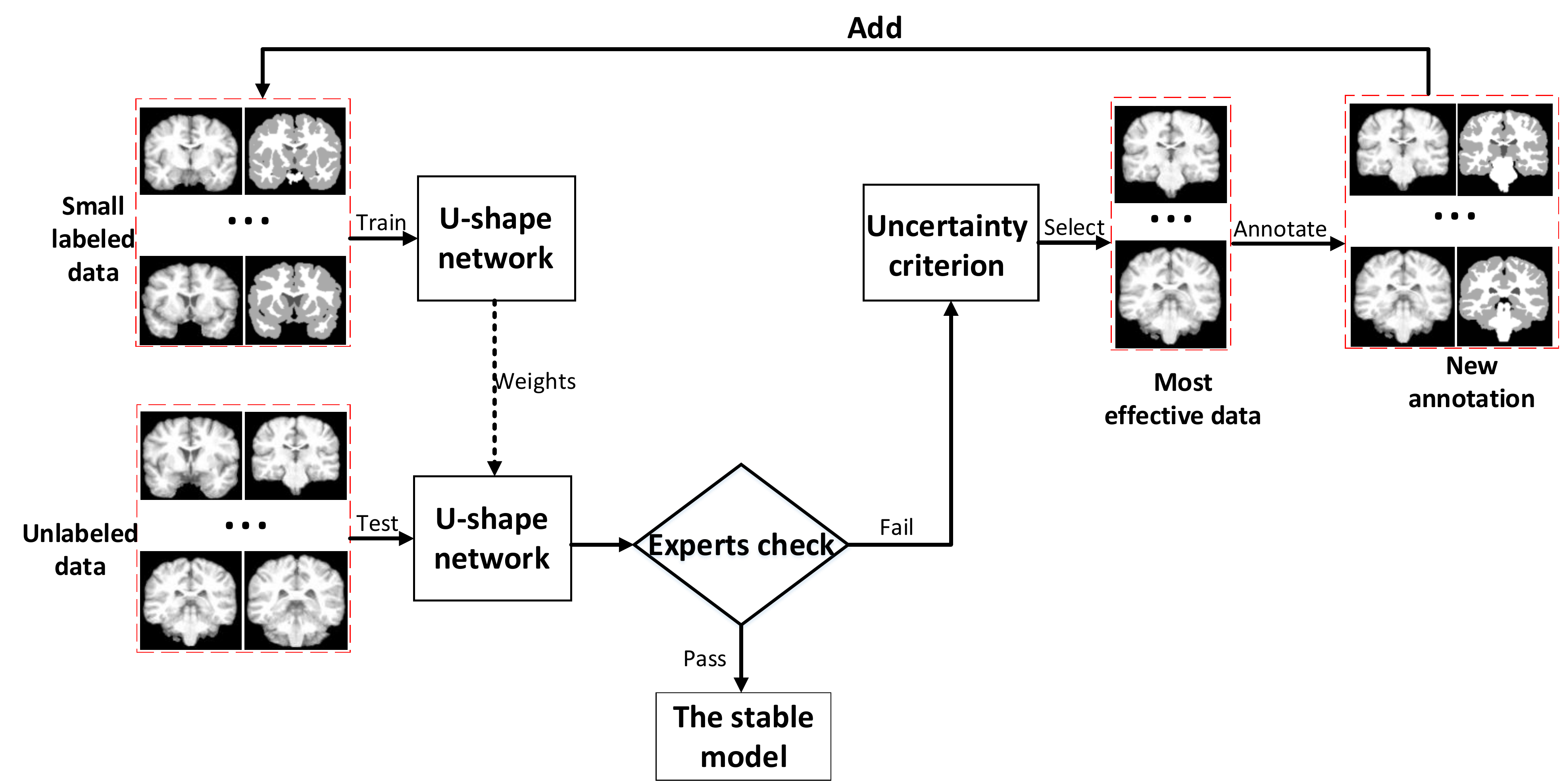}
		\caption{Flowchart of the suggestive annotation.}
	\end{center}
\end{figure}

\textbf{U-shape network}

Fig.4. shows the network architecture we use in this paper. Like the standard U-Net\cite{Ronneberger2015U}, it has an analysis and a synthesis path each with four resolution steps. In the analysis path, each layer contains two 3×3 convolutions each followed by a rectified linear unit (ReLU), and then a 2×2 max pooling with strides of 2 for down-sampling. In the synthesis path, each layer consists of an up-convolution of 2×2 by strides of one in each dimension, followed by two 3×3 convolutions each followed by a ReLU. Shortcut connections from layers of equal resolution in the analysis path provide the essential high-resolution features to the synthesis path [1]. Differing from the standard U-Net\cite{Ronneberger2015U}, in the last layer, we use 4 1×1 convolutions followed by softmax activation to reduce the subject of output channels to the subject of labels which is 4 in our case. Therefore, our network can segment CSF, GM and WM three tissues at once. At the same time, to keep the same shape after convolution, we use the same padding. The architecture has 3.1x10$^{7}$ parameters in total.

Like suggested in\cite{Szegedy2016Rethinking} we avoid bottlenecks by doubling the subject of channels already before max pooling. We also adopt this scheme in the synthesis path. The input size to the network is 64x64 and the output is 64x64x4, where the four channel separately represents the probability of the background, CSF, GM, and WM for each pixel. 

\begin{figure}[H]
	\begin{center}
		\includegraphics[width=1.\linewidth]{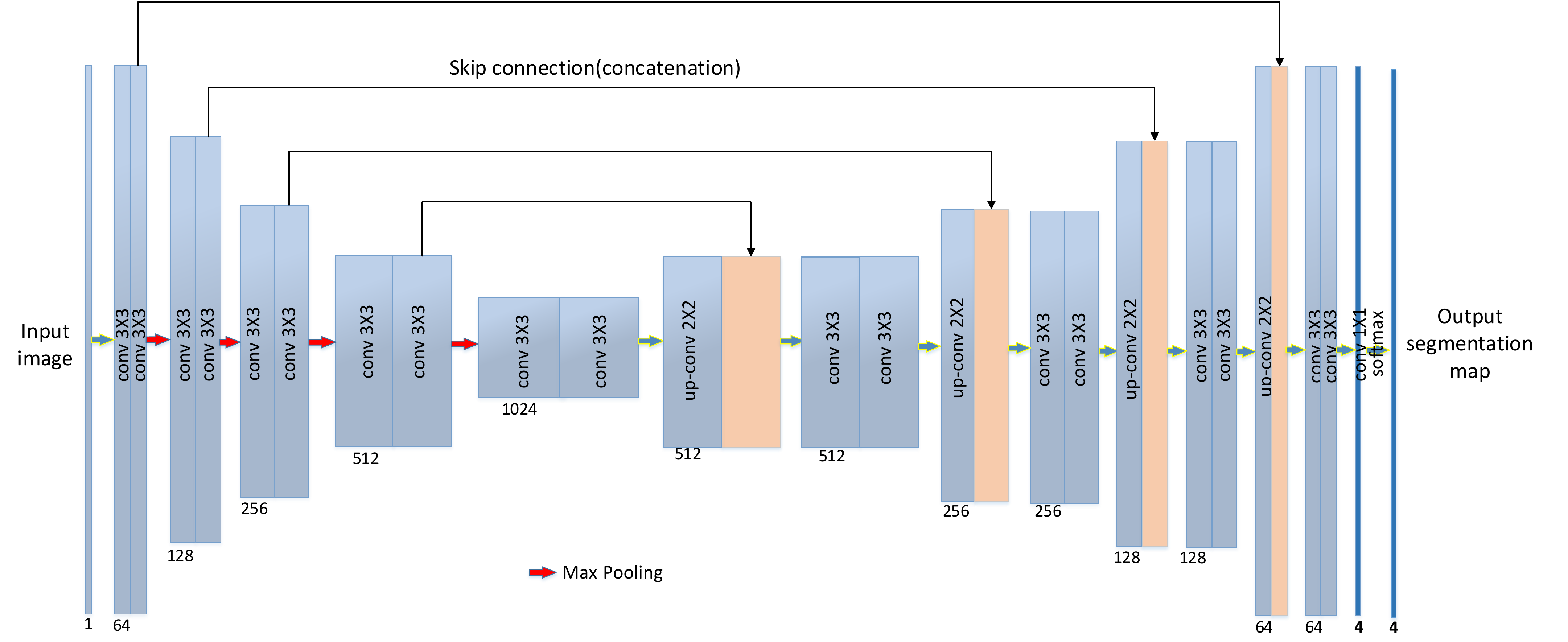}
		\caption{U-shape network architecture.}
	\end{center}
\end{figure}
Based on the goal to maximize DSC (Dice’s coefficient, the higher is better) of brain tissues, we directly use DSC loss function as following:

\begin{equation}
L(y,y')=m-\sum_{i=0}^{m-1}DSC(y_i,y_i')
\end{equation}
where $y_i$ and $y_i'$ are predicted and ground-truth for class i, respectively.Since our goal is to segment 4 classes including CSF, GM, WM and background, m is 4 here.

In the stage of segmentation reconstruction, we select the maximum probability among four classes and return the corresponding label for each pixel. In fact, we have a trail to use MSE (L2 loss) as loss function but its results are worse than using DSC loss.\\

\textbf{Uncertainty criterion}

We utilize uncertainty to determine the “worthiness” of a candidate for annotation. As mentioned below, utilizing DSC loss and softmax activation, we can attain 4 probabilities for each pixel. Referring BvSB criterion\cite{Joshi2009Multi}, which considers the difference of the highest two classes of probabilities for each pixel as a measure of uncertainty, we calculate Average BvSB defined below, because we segment a whole image at a time.

\begin{equation}
Average BvSB=\frac{\sum_{i=1}^{n}min(p(y_{Best}|x_i)-p(y_{Second-Best}|x_i))}{n}
\end{equation}
Where $p(y_{Best}|x_i)$ and $p(y_{Second-Best}|x_i)$ are the probability of pixel $x_i$ belonging to the best and the second best class, respectively, n means the total pixel number. Lower Average BvSB means higher uncertainty. Using Average BvSB can eliminate the effects of noise in image and it is easy to calculate.

\textbf{Annotation strategy}

First, we utilize very little training data to train U-shape network model and exploit it to test unlabeled data. If the test results cannot meet the requirements of experts, we use the uncertainty extracted by well-trained U-shape network to decide what will be the next data to annotate. By adding the new annotation to original training data, we iteratively train the model until performance is satisfactory. Finally, we attain a stable model which can achieve state-of-the-art performance by annotating the most effective data instead of annotating the full unlabeled data. 

\begin{figure}[H]
	\begin{center}
		\includegraphics[width=1.\linewidth]{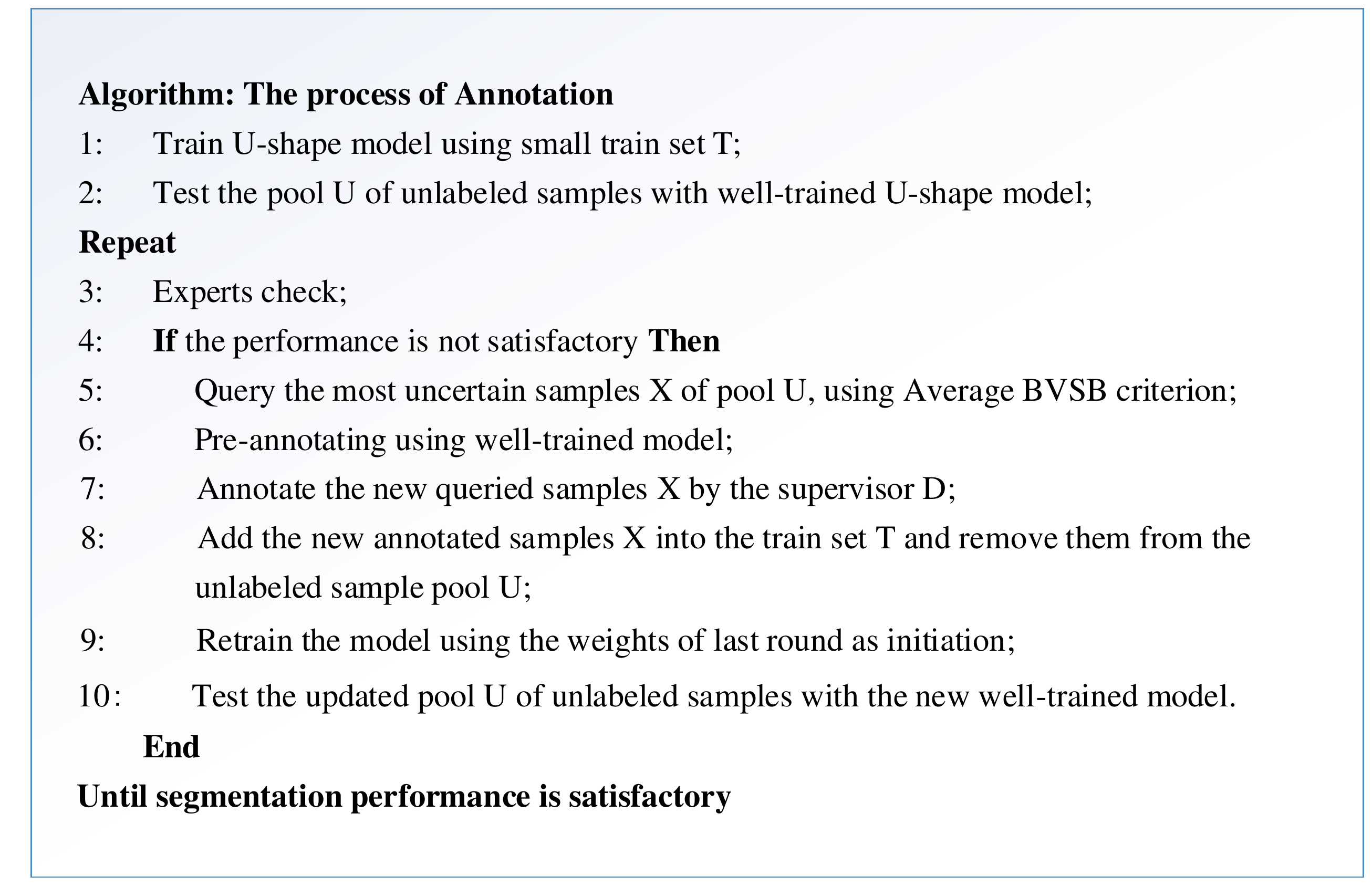}
		\caption{The process of Annotation.}
	\end{center}
\end{figure}

\subsection{Fine annotation}

In annotation stage of Fig.3, the annotator doesn’t need to annotate the most effective annotation candidate selected by active learning from scratch. They just need to correct the wrong predictions like Fig.6. 

\begin{figure}[H]
	\begin{center}
		\includegraphics[width=1.\linewidth]{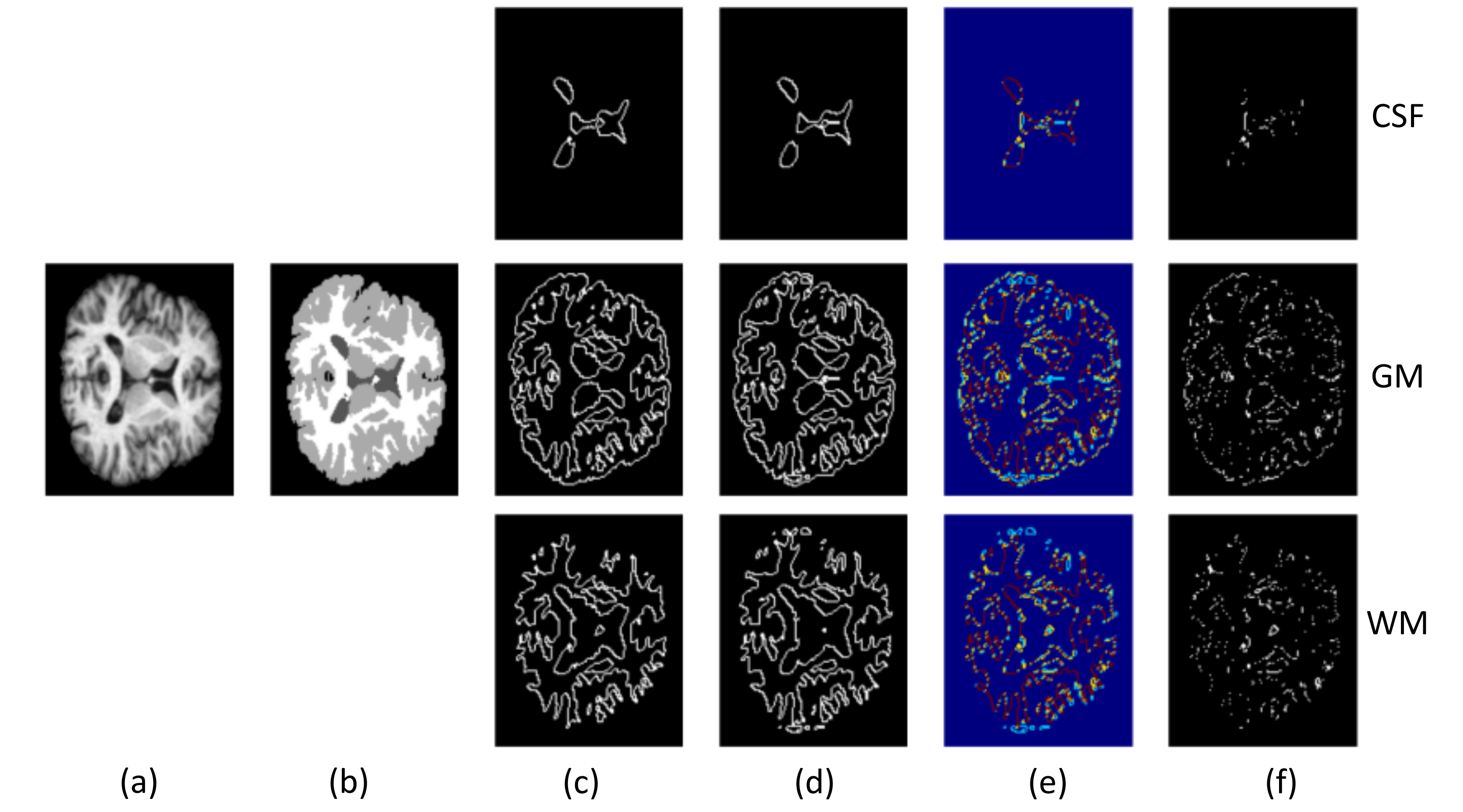}
		\caption{New annotation. (a) original images; (b) complete ground truths; (c) single class ground truths; (d) single class predictions; (e) overlap graphs of ground truth and prediction(the dull red lines mean the common part of ground truths and predictions, the blue lines mean the predictions and the yellow lines mean the ground truths); (f) extra parts needed to be annotated.}
	\end{center}
\end{figure}

As is illustrated in Fig.7, the green lines mean the predictions and the blue lines are the ground truths. The doctor should correct line ab of green instead of annotating from scratch. Therefore, the real annotation efforts should be the time spent annotating line ab. We calculate the saved efforts below.

\begin{equation}
\setlength\belowdisplayskip{-8pt}
saved \quad efforts=\frac{C}{L}\times100\%
\end{equation}
Where the length of ground truths is L and the length of overlapping part of ground truths and predictions is C.

\begin{figure}[H]
	\setlength{\belowcaptionskip}{-1cm}
	\begin{center}
		\includegraphics[width=6cm,height=5cm]{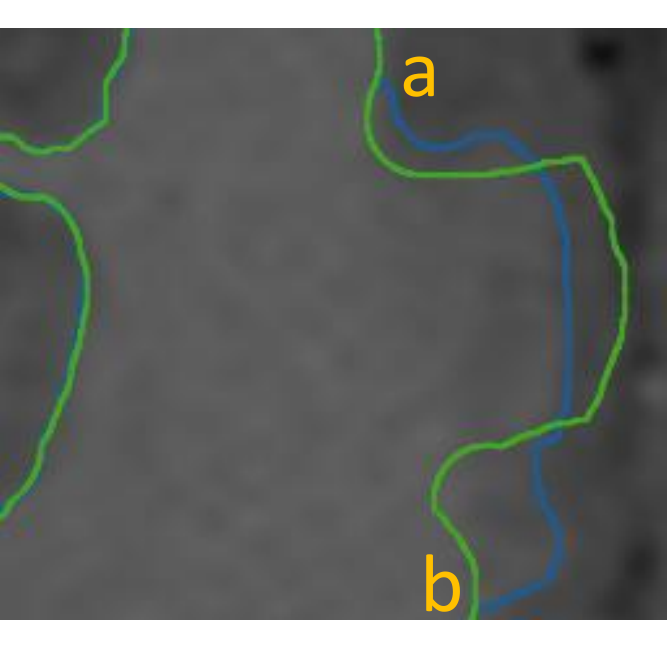}
		\caption{Diagram of calculating annotation efforts (point a and b are the intersections).}
	\end{center}
\end{figure}

\section{Experiments and Results}
\subsection{The most effective annotation candidates}
To thoroughly evaluate our method on different scenarios, we apply it to the IBSR18 dataset and MRbrainS18 Challenge dataset. The IBSR18 dataset consists of 18 T1 mode MRI volumes (01-18) and corresponding ground truth (GT) are provided. We use 10 samples as the full training data and the other 8 samples as testing data. The MRbrainS18 Challenge provides 7 labeled volumes (1, 4, 5, 7, 14, 070, 148) as training data but no testing data. We divide the 7 annotated data into two parts (5 as the full training data and 2 as testing data). We use cross 2-fold cross-validation. 

Using full training data, we compare our method with several state-of-the-art methods, including Moeskops’ multi-scale (25$^2$,51$^2$,75$^2$ pixels) patch-wise CNN method\cite{Moeskops2016Automatic} and Chen’s voxel based residual network\cite{Chen2017VoxResNet} on the two datasets. Below are the results.

\begin{table}[H]
	\centering
	\begin{tabular}{cccccc}
		\toprule
		Method & CSF & GM & WM &Average &Time(s)\\
		\midrule
		U-shape network(ours)&\textbf{89.75}&\textbf{91.18}&\textbf{91.80}&\textbf{90.91}&\textbf{40}\\
		VoxResNet\cite{Chen2017VoxResNet}& 81.03 & 87.91 & 89.73 & 86.2 & 100\\
		Multi-scale CNN\cite{Moeskops2016Automatic}& 63.01 & 80.53 & 82.16 & 75.23 & 3500\\
	    \bottomrule
	\end{tabular}
	\caption{Comparison with full training data for IBSR18 dataset segmentation(DSC: \%).}
\end{table}

\begin{table}[H]
	\centering
	\begin{tabular}{ccccc}
		\toprule
		Method& CSF& GM&WM&Average\\
		\midrule
		U-shape network(ours)&\textbf{90.78}&\textbf{83.98}&\textbf{87.45}&\textbf{87.40}\\
		VoxResNet\cite{Chen2017VoxResNet}& 90.38& 82.97&84.80&86.05\\
		\bottomrule
	\end{tabular}
	\caption{Comparison with full training data for MRBrainS18 dataset segmentation(DSC: \%).}
\end{table}

From Table 1 and Table 2, we can see that our U-shape model achieves considerable improvement on all columns. 

Then, we evaluate the effectiveness of proposed annotation strategy. We use DSC of testing data as checking standard of experts and original ground truth as new annotation in Fig.3. For IBSR18 dataset, we randomly initialize the small train set T with 2 training data and we regard the remaining 8 training data as unlabeled data, while for MRBrainS18 Challenge dataset, we randomly initialize the train set T with 1 training data and the remaining 4 training data as unlabeled data. We query the most uncertain sample X each time. After that, we add the new labeled sample X into the train set T and retrain the model until satisfying stopping criterion. To save training time, we retrain the model using the weights of last round as initiation instead of training from scratch.

\begin{figure}[H]
	\begin{center}
		\includegraphics[width=1.\linewidth]{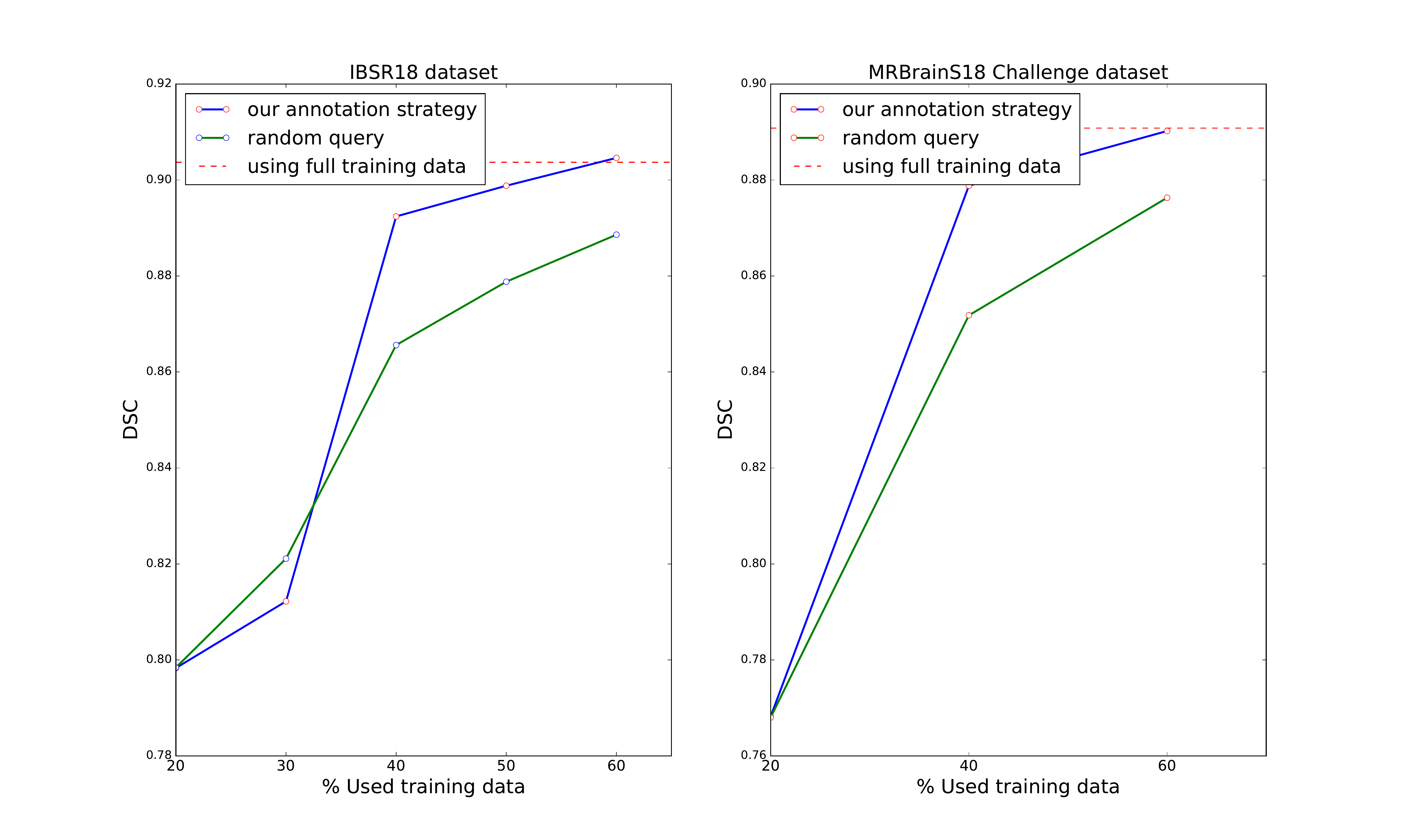}
		\caption{Comparison using limited training data for MR brain tissue segmentation.}
	\end{center}
\end{figure}

As is shown in Fig.8, we compare our method with random query: randomly requesting annotation. It shows that our annotation strategy is consistently better than random query, and state-of-the-art performance can be achieved by using only 60\% of the training data.

\subsection{Efforts evaluation on each annotation candidate}

As mentioned in 3.1, we randomly initialize the small train set T with 2 training data on IBSR18 dataset and iteratively train U-shape network to select the most effective one sample for annotation. For MRBrainS18 Challenge dataset, starting with a completely empty labeled dataset, we use transfer learning to attain the first training sample. Using annotation platform in 2.2, the annotator just need to correct wrong predictions. Below are the results of saved annotation efforts.

\begin{table}[H]
	\centering
	\begin{tabular}{cccccc}
		\toprule
		Class& Iteration1& Iteration2&Iteration3&Iteration4&mean\\
		\midrule
		CSF&73.04&47.96&49.56&46.82&54.35\\
    	GM& 63.38& 39.68&43.57&41.58&47.05\\
		WM& 62.80& 39.57&43.24&43.13&47.19\\
		\bottomrule
	\end{tabular}
	\caption{Saved efforts (\%) on IBSR18 dataset.}
\end{table}

\begin{table}[H]
	\centering
	\begin{tabular}{ccccc}
		\toprule
		Class& Iteration1& Iteration2&Iteration3&mean\\
		\midrule
		CSF&2.46&56.28&73.71&44.15\\
		GM& 34.87& 42.25&56.83&44.65\\
		WM& 48.79& 49.03&49.31&49.04\\
		\bottomrule
	\end{tabular}
	\caption{Saved efforts (\%) on MRBrainS18 Challenge dataset.}
\end{table}

\subsection{Implementation details and computation cost}
In our work, when using the full training data, the network was trained for 500 epochs on a single NVIDIA TitanX GPU. In order to prevent the network from over-fitting, we applied early stopping in the training process. The training process was automatically terminated when the validation accuracy did not increase after 30 epochs in Fig.9, which took approximately 3 hours for the whole training process. We used the glorot\_uniform initialization and the Adam algorithm in keras. Segmentation runtime is 40-50 seconds for processing each testing data (size 256 ×128×256). In retraining process, training time was about 1 hours due to the using of initiation.

\begin{figure}[H]
	\begin{center}
		\includegraphics[width=8cm,height=6cm]{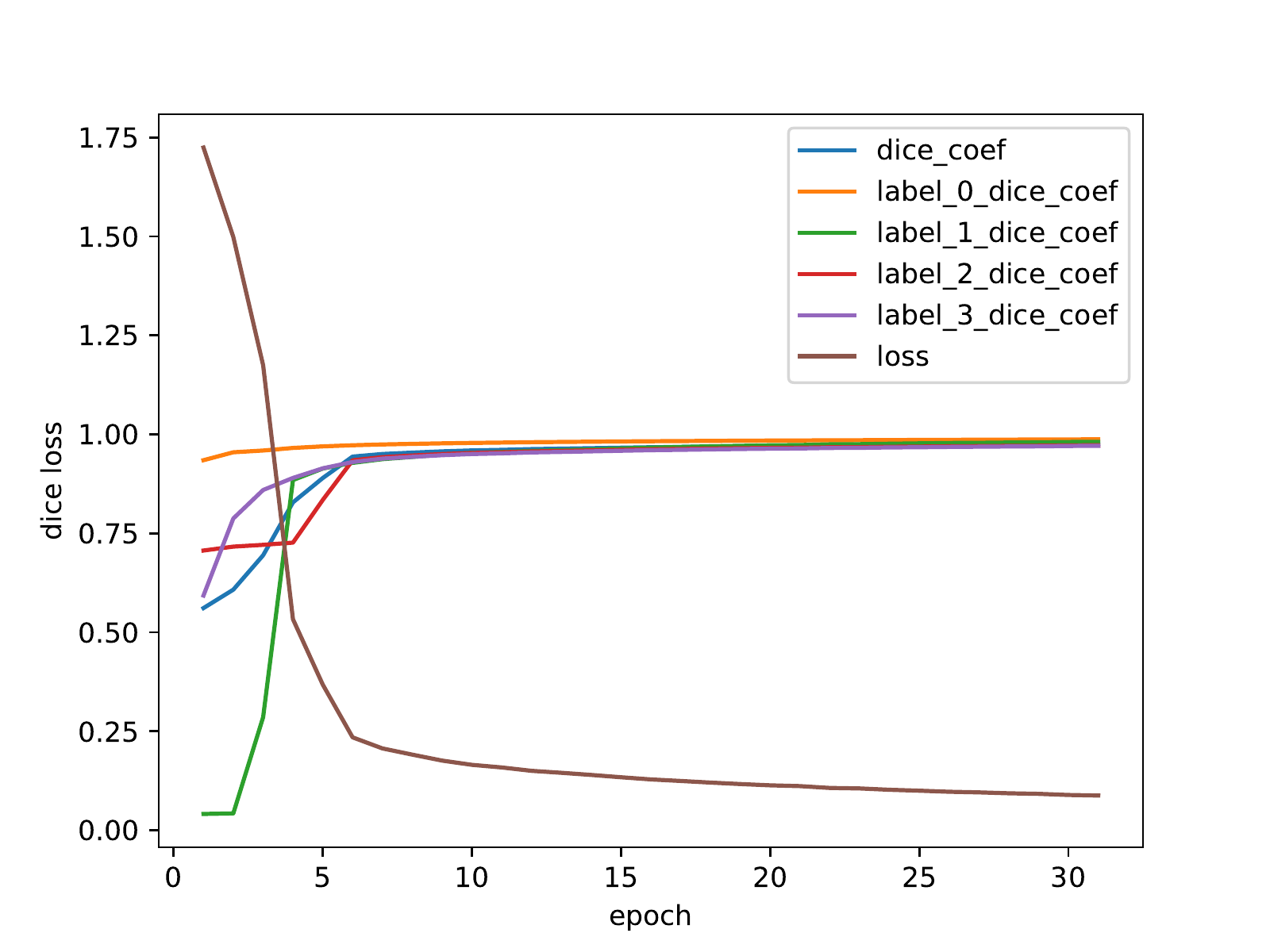}
		\caption{Dice loss as a function of a training epoch for our proposed models.(label\_0\_dice-coef, label\_1\_dice-coef , label\_2\_dice-coef , label\_3\_dice-coef means the DSC of background, CSF, GM and WM respectively.)}
	\end{center}
\end{figure}

\section{Discusion}
By comparing the segmentation results illustrated in Fig.10, we realize that different models have what they are severally skilled in. Although the U-shape network model performs better than VoxResNet method does in general, as the red boxes show, the green box indicates that VoxResNet model also has its advantages.

\begin{figure}[H]
	\begin{center}
		\includegraphics[width=1.\linewidth]{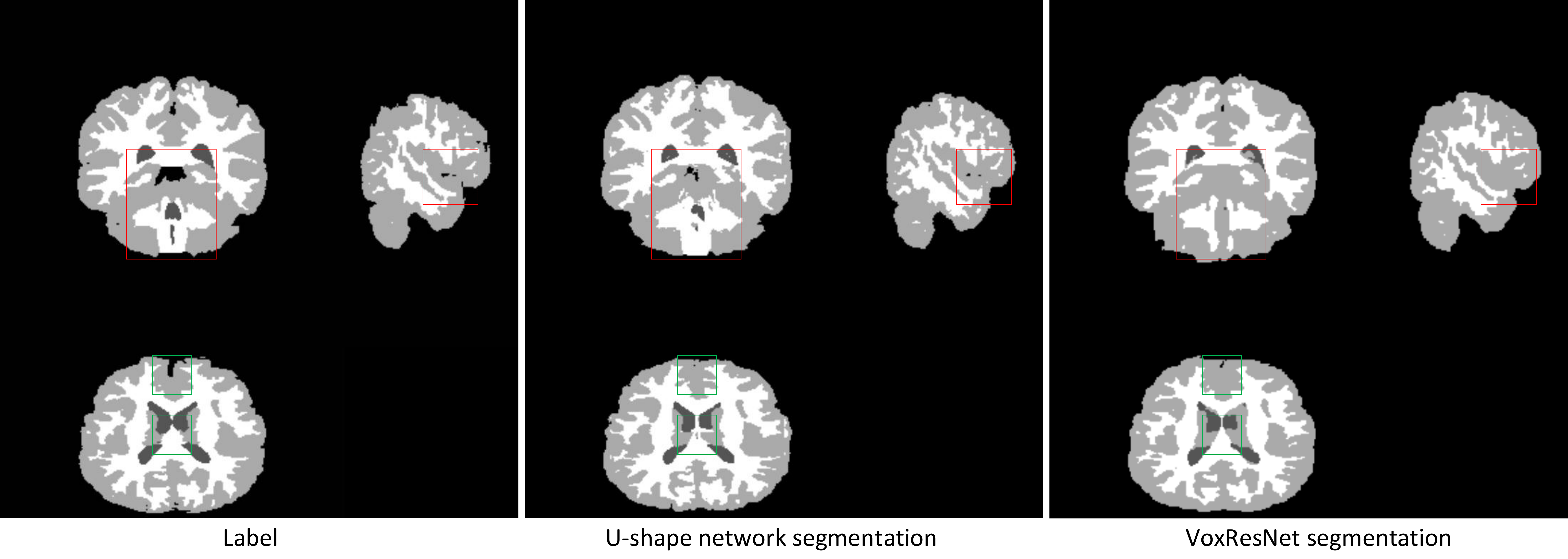}
		\caption{Visual results of different methods.}
	\end{center}
\end{figure}

To explore the relationship between data and deep learning models, we combine the strategy described in section 2.3 with VoxResNet model. We find that the most effective training data selected by VoxResNet model differ from those selected by U-shape network model. For U-shape network model, start-of-the-art performance can be achieved by using 03, 06, 10-13 of IBSR18 dataset as training data, while for VoxResNet model, the most effective training data are 01, 02, 10-14. Meanwhile, for MRBrainS18 dataset, the most effective data for U-shape network model are 1, 4, 148 while they are 1, 5, 7 for VoxResNet model. In other words, “worthiness” differs from model to model. As the saying goes, a foot is short and an inch is long. Therefore, it is a promising way to combine the advantages of different models, which is what we will continue to research in our future studies.

In this paper, we conduct experiments on brain MR data, which is difficult to annotate due to the high complex structure and little grayscale change between different tissue classes. We need to evaluate proposed method on more types of data and more clinical data in the future. Compared with MR, CT image has the advantages of high density resolution and is extensively used in radiotherapy. However, the doctor always needs to spend much time in delineation before making radiotherapy plan. Therefore, our proposed method is expected to dramatically alleviate burden in radiotherapy. Meanwhile, besides brain, our proposed method can be used in other positions like chest and abdominal CT.

\section{Conclusions}
In this paper, we propose an efforts estimation of doctors annotating medical image.  There are two main contributions.
(1) An effective suggestive annotation strategy to select the most effective training data, which can attain state-of-the-art performance by using only 60\% training data; (2) An annotation platform to alleviate annotation efforts on each selected effective annotation candidate, which can cut at least 44\%, 44\%, 47\% annotation efforts.

\bibliographystyle{IEEEtranS}
\bibliography{3D-DRL}

\begin{thebibliography}{10}
\providecommand{\url}[1]{#1}
\csname url@samestyle\endcsname
\providecommand{\newblock}{\relax}
\providecommand{\bibinfo}[2]{#2}
\providecommand{\BIBentrySTDinterwordspacing}{\spaceskip=0pt\relax}
\providecommand{\BIBentryALTinterwordstretchfactor}{4}
\providecommand{\BIBentryALTinterwordspacing}{\spaceskip=\fontdimen2\font plus
\BIBentryALTinterwordstretchfactor\fontdimen3\font minus
  \fontdimen4\font\relax}
\providecommand{\BIBforeignlanguage}[2]{{%
\expandafter\ifx\csname l@#1\endcsname\relax
\typeout{** WARNING: IEEEtranS.bst: No hyphenation pattern has been}%
\typeout{** loaded for the language `#1'. Using the pattern for}%
\typeout{** the default language instead.}%
\else
\language=\csname l@#1\endcsname
\fi
#2}}
\providecommand{\BIBdecl}{\relax}
\BIBdecl

\bibitem{google}
F.~V. Andriluka~M, Uijlings J R~R, ``Fluid annotation: A human-machine
  collaboration interface for full image annotation,'' 2018.

\bibitem{Bao2015Multi}
S.~Bao and A.~C.~S. Chung, ``Multi-scale structured cnn with label consistency
  for brain mr image segmentation,'' vol.~6, no.~1, pp. 1--5, 2015.

\bibitem{Chen2017VoxResNet}
H.~Chen, Q.~Dou, L.~Yu, J.~Qin, and P.~A. Heng, ``Voxresnet: Deep voxelwise
  residual networks for brain segmentation from 3d mr images,''
  \emph{Neuroimage}, vol. 170, p. S1053811917303348, 2017.

\bibitem{Dong2016Fully}
N.~Dong, W.~Li, Y.~Gao, and D.~Sken, ``Fully convolutional networks for
  multi-modality isointense infant brain image segmentation,'' in \emph{IEEE
  International Symposium on Biomedical Imaging}, 2016.

\bibitem{Hao2016Deep}
C.~Hao, X.~Qi, J.~Z. Cheng, and P.~A. Heng, ``Deep contextual networks for
  neuronal structure segmentation,'' in \emph{Thirtieth Aaai Conference on
  Artificial Intelligence}, 2016.

\bibitem{Hao2016DCAN}
C.~Hao, X.~Qi, L.~Yu, and P.~A. Heng, ``Dcan: Deep contour-aware networks for
  accurate gland segmentation,'' in \emph{Computer Vision \& Pattern
  Recognition}, 2016.

\bibitem{Hong2015Decoupled}
S.~Hong, H.~Noh, and B.~Han, ``Decoupled deep neural network for
  semi-supervised semantic segmentation,'' 2015.

\bibitem{Jain2016Active}
S.~D. Jain and K.~Grauman, ``Active image segmentation propagation,'' 2016.

\bibitem{Joshi2009Multi}
A.~J. Joshi, F.~Porikli, and N.~Papanikolopoulos, ``Multi-class active learning
  for image classification,'' in \emph{IEEE Conference on Computer Vision \&
  Pattern Recognition}, 2009.

\bibitem{Moeskops2016Automatic}
P.~Moeskops, M.~A. Viergever, A.~M. Mendrik, L.~S.~D. Vries, M.~J. N.~L.
  Benders, and I.~Isgum, ``Automatic segmentation of mr brain images with a
  convolutional neural network,'' \emph{IEEE Transactions on Medical Imaging},
  vol.~35, no.~5, pp. 1252--1261, 2016.

\bibitem{Ronneberger2015U}
O.~Ronneberger, P.~Fischer, and T.~Brox, \emph{U-Net: Convolutional Networks
  for Biomedical Image Segmentation}.\hskip 1em plus 0.5em minus 0.4em\relax
  Springer International Publishing, 2015.

\bibitem{Szegedy2016Rethinking}
C.~Szegedy, V.~Vanhoucke, S.~Ioffe, J.~Shlens, and Z.~Wojna, ``Rethinking the
  inception architecture for computer vision,'' in \emph{Computer Vision \&
  Pattern Recognition}, 2016.

\bibitem{Wang2016A}
Z.~Wang, D.~Bo, L.~Zhang, and L.~Zhang, ``A batch-mode active learning
  framework by querying discriminative and representative samples for
  hyperspectral image classification ¡î,'' \emph{Neurocomputing}, vol. 179,
  no.~C, pp. 88--100, 2016.

\bibitem{Xu2016Gland}
Y.~Xu, Y.~Li, M.~Liu, Y.~Wang, M.~Lai, and I.~C. Chang, ``Gland instance
  segmentation by deep multichannel side supervision,'' \emph{IEEE Transactions
  on Biomedical Engineering}, vol.~PP, no.~99, pp. 1--1, 2016.

\bibitem{Yang}
L.~Yang, Y.~Zhang, J.~Chen, S.~Zhang, and D.~Z. Chen, ``Suggestive annotation:
  A deep active learning framework for biomedical image segmentation,'' in
  \emph{Medical Image Computing and Computer-Assisted Intervention ? MICCAI
  2017}.

\bibitem{Zhang2015Deep}
W.~Zhang, R.~Li, H.~Deng, L.~Wang, W.~Lin, S.~Ji, and D.~Shen, ``Deep
  convolutional neural networks for multi-modality isointense infant brain
  image segmentation,'' \emph{Proc IEEE Int Symp Biomed Imaging}, vol. 108, pp.
  1342--1345, 2015.

\bibitem{Zhou2017Fine}
Z.~Zhou, J.~Shin, Z.~Lei, S.~Gurudu, M.~Gotway, and J.~Liang, ``Fine-tuning
  convolutional neural networks for biomedical image analysis: Actively and
  incrementally *,'' in \emph{IEEE Conference on Computer Vision \& Pattern
  Recognition}, 2017.

\end{thebibliography}

\end{spacing}
\end{document}